\documentclass[12pt, anon]{l4dc2022}

\usepackage{mdframed}

\newmdtheoremenv{problem}{Optimization}
\newcommand{\norm}[1]{\left\lVert#1\right\rVert}

\title{Accelerating Trajectory Generation for Quadrotors Using Transformers}

\author{%
 \Name{Srinath Tankasala} \Email{stankasa@utexas.edu}\\
 \addr The University of Texas at Austin
 \AND
 \Name{Mitch Pryor} \Email{mpryor@utexas.edu}\\
 \addr The University of Texas at Austin%
}

\begin{document}

\maketitle

\begin{abstract}%
 In this work, we address the problem of computation time for trajectory generation in quadrotors. Most trajectory generation methods for waypoint navigation of quadrotors, for example minimum snap/jerk and minimum-time, are structured as bi-level optimizations. The first level involves allocating time across all input waypoints and the second step is to minimize the snap/jerk of the trajectory under that time allocation. Such an optimization can be computationally expensive to solve. In our approach we treat trajectory generation as a supervised learning problem between a sequential set of inputs and outputs. We adapt a transformer model to learn the optimal time allocations for a given set of input waypoints, thus making it into a single step optimization. We demonstrate the performance of the transformer model by training it to predict the time allocations for a minimum snap trajectory generator. The trained transformer model is able to predict accurate time allocations with fewer data samples and smaller model size, compared to a feedforward network (FFN), demonstrating that it is able to model the sequential nature of the waypoint navigation problem. 
\end{abstract}


\section{Introduction}
	
    A quadrotor is an extremely agile aerial system that has found widespread use in commercial activities such as inspections, package delivery, cinematography, etc. Multi-waypoint trajectory generation is essential for autonomous systems navigation and is typically done using a local planner. Local planners for quadrotor navigation typically minimize smooth polynomial trajectories based on aggregate snap/jerk. The most widely used formulation in minimum snap/jerk planners is a bi-level optimization where the time allocations across waypoints is computed in the first level and a constrained optimization of polynomial coefficients over different waypoints forms the second level \cite{mellinger2011minimum, richter2016polynomial}. This strategy of optimizing piecewise time allocations and then optimizing piecewise polynomials between waypoints is also used in minimum-time trajectory generation for drones \cite{zhou2019robust, gao2018online}. 
    
    For minimum snap trajectory generation, recent works \cite{burke2021fast, wang2021generating} have cleverly reformulated both steps of the bi-level optimizations that are solvable in linear time. In this work, we propose to instead learn the optimal waypoint time allocations using deep learning models. By eliminating the bi-level structure of the optimization problem, we can accelerate the trajectory generation process. This would require the learning-based model to be capable of giving highly accurate predictions of optimal time allocations for any given sequence of waypoints. In our approach we use sequence models, specifically transformers\cite{vaswani2017attention}, to learn the optimal time allocations for minimum snap trajectory generation. While the approach here is used to learn parameters in the two level minimum snap optimization, this approach can be extended to learn parameters for other trajectory generation problems such as minimum-time trajectories. 
    
    This work has three main contributions: (1) we use transformers to learn optimization parameters that removes the bi-level structure and turns it into a single-level optimization problem making it faster (2) we evaluate the trained transformer model performance, sample efficiency and also show its effectiveness across different settings, including input sizes greater than those of samples in the training data (3) we study the learnt attention patterns to understand the input output relation learnt by the model and the insight it provides about the nature of the time allocation optimization. Overall, this seq2seq modeling approach can be used to learn parameters from offline trajectory data, and can provide good approximations to the time allocations thus speeding up trajectory optimization.


\section{Related Work}

The differential flatness property of quadrotor dynamics is well known in the literature (\cite{mellinger2011minimum}) and it allows the user to plan trajectories in the flattened output space of the quadrotor, i.e. position, velocity, acceleration and yaw. Notable approaches for trajectory generation in quadrotors use sequential convex optimization \cite{hargraves1987direct} and sampling based kinodynamic planning \cite{lavalle2001randomized} among others. The most widely used trajectory generation methods for waypoint navigation, however, use polynomial functions as basis/primitives for generating their trajectories\cite{mellinger2011minimum, richter2016polynomial}. 

The bi-level optimization structure is also used in minimum-time trajectory generation in the non-linear quadrotor state space \cite{zhou2019robust, tankasala2022, tankasala2022smooth}. However, allocating the total time across different waypoints requires iterative refinement which increases computation time. Heuristic functions have been used \cite{gao2018online} to speed up this step, but they can lead to suboptimal solutions. Alternative formulations, such as \cite{foehn2021time}, circumvent the time allocation problem by reformulating the optimization with complementary progress constraints (CPC). However, the method in \cite{foehn2021time} is still computationally expensive. 

Thus in many applications, quadrotor trajectory planning is typically formulated as a bi-level optimization problem with the first level involving time allocations across desired waypoints. For minimum jerk/snap optimizations, the time allocation problem can be solved using a backtracking gradient descent (BGD) until convergence to a local minimum \cite{mellinger2011minimum, liu2016high}. These methods typically use a starting guess for the time allocation such as a trapezoidal velocity profile (TVP) \cite{liu2016high}. This time allocation optimization step has been computationally expensive, until recently where \cite{burke2021fast} and \cite{wang2021generating} have solved it in linear time.

An alternative approach that has gained traction is the use of data driven methods \cite{ryou2020multi, de2019real}. While supervised learning models like MLPs (multi-layer perceptron) can be trained to predict the waypoint time allocations \cite{de2019real}, they cannot handle variable number of input waypoints and hence are not usable beyond a certain input size. In \cite{ryou2020multi}, the authors train a deep Gaussian Process (GP) to evaluate the feasibility of waypoint time allocations and refine them. In later work \cite{ryou2022realtime}, the authors train a VAE(variational autoencoder) to guess optimal time allocations. They then train a PPO policy to refine those time allocations for minimum time trajectory generation. 

Given that the waypoint trajectory generation problem is sequential in nature, all input waypoints need to be considered when predicting time allocation for a given segment between waypoints. Hence traditional RNNs cannot be used as they do not consider future inputs when making predictions at a given step in the input sequence. We use transformers \cite{vaswani2017attention} as they consider all inputs simultaneously when generating the time allocations. Their attention function implementations have also been optimized \cite{wang2020linformer} to make them computationally fast. Hence they are suitable for learning the time allocations over the given input waypoints. 


\section{Approach}
In minimum snap planning, we optimize the aggregate snap over the entire trajectory. We denote the 3D trajectory of the UAV as $r(t) = [x(t), y(t), z(t)]$. The minimum-snap trajectory optimization is solved separately along each Cartesian axis (x,y,z), described in \cite{mellinger2011minimum}, and it is formulated as: 
\begin{align}
& \quad \quad \min_{r,\underline{t}}  J(r,\underline{t}) \notag\\
&\text{s.t. } \ \ r(0) = w_0; \quad v(0) = v(T) = 0 \label{eq:snap_optim1}\\
& \quad \quad r_k(\sum_{j=1}^i t_j) = w_i; \quad \Sigma t_i = T; \ \forall i=1,2,...,m \notag\\
& \text{where, }\quad J(r,\underline{t}) = \int_{0}^{T}\norm{d^4{\rm r}\over dt^{4}}_2^{2} dt; \quad r\in R_T \label{eq:min_snap_obj}
\end{align}
$\underline{t} = [t_1,\ t_2,...,\ t_m]$, is the allocations of total time $T$ over the desired waypoints $W = [w_0, w_1,\ w_2,...,\ w_m],\ w_i\in \mathbb{R}^3$. $R_T$ is the set of all feasible trajectories passing through $W$ that can be tracked in time $T$. In \cite{mellinger2011minimum}, they also include second derivative of yaw in the cost $J(r,\underline{t})$, but in many applications the yaw is planned separately or set to a constant. Hence we do not include yaw as an optimization parameter.

To solve for the minimum snap trajectory optimization, we first reduce the aggregate trajectory snap along each axis for a given $\underline{t}$. The solution is of the form of $m$ polynomials of order $n$ (7 in this case) as shown in eq \eqref{eq:rstart}. 
\begin{equation}
    r_k(t) = 
 \begin{cases}
\sum_{j=0}^{n} a_{j,1}t^j,\ \ \ t_0\leq t \leq t_1\\
 \vdots\\
\sum_{j=0}^{n} a_{j,m}t^j,\ \ \ t_{m-1}\leq t \leq t_m
 \end{cases}
 \label{eq:rstart}
\end{equation}
Given a time allocation $\underline{t}$, the minimum snap trajectory under $\underline{t}$ is a quadratic program given by Optimization \ref{prb: min_snap2}, refer \cite{mellinger2011minimum}:
\begin{problem}\label{prb: min_snap2}
\begin{align}
    \min_{a}\quad &a^T Q a \notag\\
    s.t.\quad & A_{eq} a = b_{eq} \label{eq:min snap QP}\\
        & A_{ineq} a \leq b_{ineq} \notag
\end{align}
\end{problem}
where $Q\in R^{nm\times nm}$ is a block diagonal matrix containing powers of $t$, $A_{eq}$ is constructed from Q and the equality constraint enforces the condition of passing through waypoints and continuity of velocities and accelerations. The $A_{ineq}$ and $b_{ineq}$ are typically used to enforce corridor constraints.
To solve for $\underline{t}^*$, we iteratively refine our initial guess for the time allocation. The refinement of $\underline{t}$ is also an optimization, shown in Optimization \ref{prb:min_snap1}:
\begin{problem}\label{prb:min_snap1}
\begin{align}
    \min_{\underline{t}\epsilon R^m} J(p(\underline{t},W), T) \notag\\
    \text{s.t.} \quad \sum_{i=1}^m t_i = T
    \label{eq:snap_optim2}
\end{align}
\end{problem}
where, $p(\underline{t},W)$ is the optimal trajectory across the waypoints $w_i \in W$ obtained by solving Optimization \ref{prb: min_snap2} using time allocation $\underline{t}$. This optimization is typically done by a gradient descent on $\underline{t}$ (see \cite{mellinger2011minimum}). 

Learning the optimal time allocations, $\underline{t}^*$, can greatly accelerate the trajectory generation process as it is the most computationally intensive step in many scenarios\cite{mellinger2011minimum, zhou2019robust}. 

We assume that the total flight time $T$ is known. If the total time needs to be minimized, then we scale  $T\rightarrow \eta T,\ \eta>0$, along with the optimal allocation $\underline{t}^*\rightarrow \eta \underline{t}^*$, until no feasible trajectory $p(\eta\underline{t}^*,W)$  exists (exceeds max velocity or max acceleration constraints for example). This is typically done using a quick line search. 

\subsection{Modeling methodology}

We use the standard transformer model, \cite{vaswani2017attention}, for learning waypoint time allocations. We consider 2D trajectories from the SIGN \cite{signondb} and ILGDB \cite{ilgdb} hand drawing datasets here, but this approach can easily be extended to learn 3D trajectories. We modify the inputs and position function to suit the trajectory generation problem. We normalize the input data to make them invariant to homogeneous transforms (rotation and scaling). This is reasonable as the optimal time allocation ($\underline{t}^*$) scales linearly with total time $T$ and is also independent of the distance scale between waypoints, \cite{de2019real}. 

Normalizing the inputs reduces the amount of data needed for training as the model does not have to explicitly learn that the time allocations are invariant to homogeneous transforms, total time and scale of the trajectory.

\subsection{ Data Preparation }
\label{ssec:data_prep}
In 2D, a rotation and distance scale invariant representation of the waypoints, $W$, can be obtained using the range-angle parametrization. The range and angle vectors $d, \theta$  are given by:
\begin{align}
    \underline{d} = [l_1, l_2, ..., l_{M-1}];\quad
    \theta = [0, \theta_2, \theta_3,..., \theta_{M-1}]
\end{align}
where $d_i$ is the euclidean distance between two consecutive waypoints, i.e. $d_i = ||w_{i+1} - w_i||$. While  $\theta_i$ is the angle between consecutive waypoints, i.e. $\theta_i = \angle{(V_{i-1}, V_i)}$, where the vector $V_i = w_i - w_{i-1}$. It should be noted that $-\pi < \theta_i \leq \pi$. In 3D, the angles, $\theta_i$'s can be replaced with quaternion rotation vectors. We further non-dimensionalize the range vector $\underline{d}$, by scaling all components by a common factor.
\begin{align*}
    &\underline{d}'= \frac{1}{s}\cdot\underline{d},
\end{align*}
where, $s>0$,and $0 < d'_i \leq 1$. This reparametrization gives a representation of the input waypoints that is invariant to homogeneous transformations and is valid as the output $\underline{t}^*$ is unchanged by these transformations. We scale down the optimal time fractions ($\underline{t}^*,\  \Sigma t_i^* = T$) by the total time $T$ to obtain the optimal time allocations of the trajectory. During inference, we scale back the model output $\underline{t}^*$ to match the desired total time $T$. 

\begin{figure}[h]
    \centering
    \includegraphics[width=4.5in]{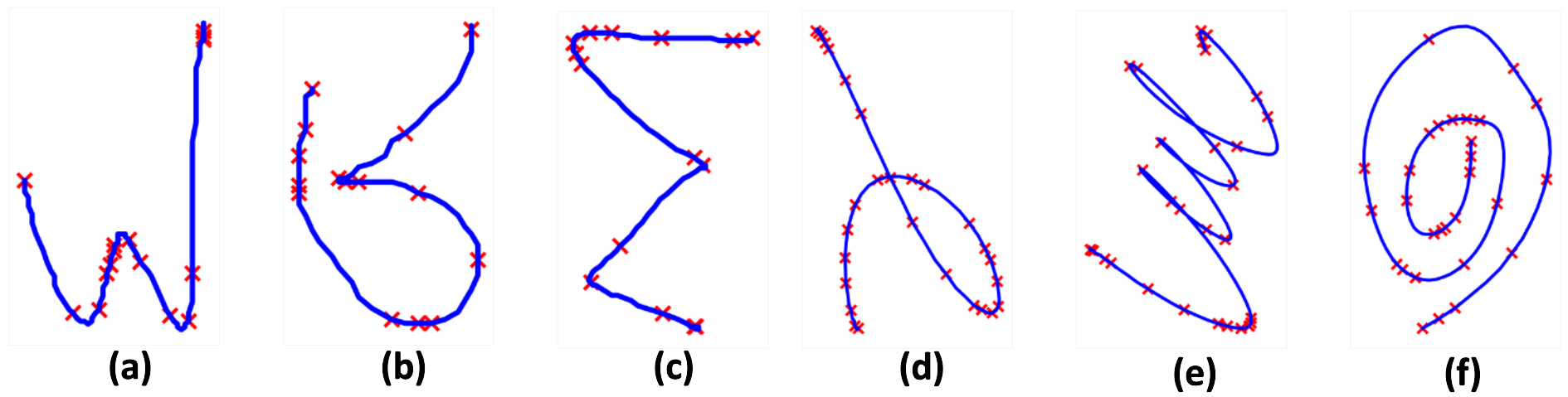}
    \caption{Sample curves collocated to 30 points from the ILGDB dataset ((a) to (c)) and SIGN dataset ((d) to (f))}
    \label{fig:sample_curves}
\end{figure}

To obtain a sample trajectory with a desired number of waypoints ($N$), we use gestures/curves taken from the handwriting curves dataset (SIGN, ILGDB). For a sample gesture/curve, we preserve the start and end points and arbitrarily collocate the rest of the curve. This provides a sample set of waypoints with the desired length, $N$. Sample sets of waypoints from the training and test dataset along with their minimum-snap trajectories  are shown in Fig \ref{fig:sample_curves}

\subsection{Transformer Model}
\label{subsec: model}
The encoder and decoder of the standard transformer \cite{vaswani2017attention} are adapted to learn the time allocation $\underline{t}$ given the input waypoints. The encoder and decoder consist of multi-head self-attention blocks and a position-wise feedforward network (FFN) block. 

\subsubsection*{Input encoding and positional encoding}

We use a linear layer to encode sample inputs ($W^{(i)}$) and outputs $T^{(i)}$ into the desired embedding space.

$W^{(i)} = [ [l_1, 0], [l_2, \theta_2], [l_3, \theta_3],...,[l_{M-1}, \theta_{M-1}] ]$ 
where $W^{(i)}$ is a single sample set of waypoints taken from training dataset. We then solve for the optimal time allocation vector based on minimum snap, using \cite{wang2021generating}, to get $T^{(i)} = [t_1, ..., t_{M-1}]$. We use the standard positional encoding function from \cite{vaswani2017attention}. The position encoding simply reflects the order in which the waypoints are visited. 

\subsubsection*{Multi-head Attention function}

The queries and keys used for our case are the encodings learnt by the input embedding layers. We use a linear layer that takes the range-angle parameters of the desired waypoints and calculates an embedding of desired size that can be used as the querys, keys and values. 

Similarly, the decoder uses a linear layer that converts the time allocations ($0,t_1,t_2,..., t_{N-1}$) to an appropriate output embedding of the desired size. The time allocation is padded with a zero in the beginning to serve as a starting point for the decoding process. Since we need only $N-1$ time allocations, the decoder is run for that many steps to get the predicted time allocation vector $\underline{t}_{pred}$. We use the same attention masks for the inputs and outputs as done in \cite{vaswani2017attention}.

\subsubsection*{Feedforward network}

The output head of the transformer is a fully connected Linear layer with a 1 dimensional output (time allocation $t_i$). We use a cumulative L1 loss between the time allocations predicted by the decoder $\underline{t}_{pred}$ and the true time allocation $\underline{t}^*$. The softmax labels layer from the standard transformer is not used as this is not a classification task, i.e. it is a regression problem. Again for convenience, we normalize the predicted time steps from the decoder $\underline{t}_{pred}$ to sum to 1. Hence the final output is given by eq \eqref{eq:tout}

\begin{equation}
    \underline{t}_{out} = \frac{\underline{t}_{pred}}{||\underline{t}_{pred}||_1} \label{eq:tout}
\end{equation}

\begin{figure}[h]
    \centering
    \includegraphics[width=2.7in]{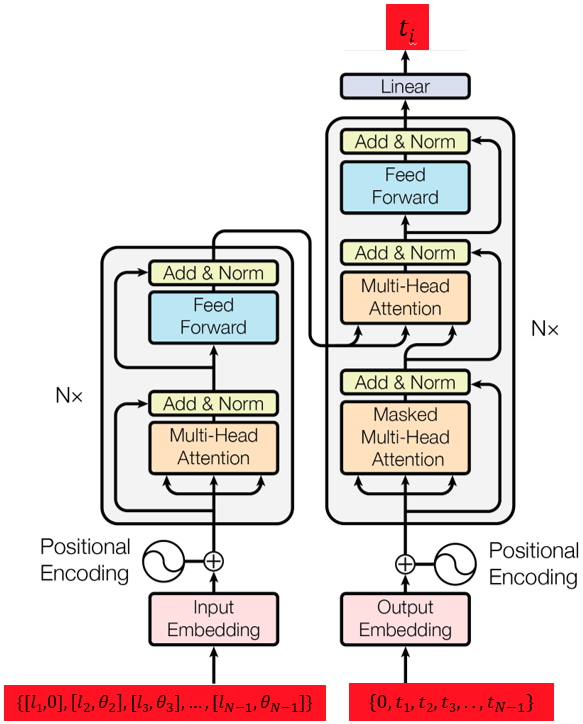}
    \caption{Modified inputs and outputs to standard transformer \cite{vaswani2017attention} (shown in red) for trajectory generation}
    \label{fig:transformer}
\end{figure}

The modifications made to the standard transformer for learning time allocations are shown in Fig \ref{fig:transformer}.

\section{Results and Analysis}
\label{sec:result}
We trained a transformer model using gestures generated from SIGN dataset as described in Section \ref{ssec:data_prep}. We use an embedding dimension of 32 for the inputs and outputs along with 16 attention heads. We use a total of 3 encoder and 3 decoder layers with a feedforward network hidden dimension of 256 in the transformer model. This results in a total of $139,329$ trainable hyperparameters. To generate the training dataset we collocated every gesture in the SIGN dataset to various resolutions, from 3 to 30 waypoints. This resulted in a total of $493,668$ sets of waypoints. We split the dataset in the ratio of 5:1 for training and validation respectively. For producing the test dataset we collocated every curve in the ILGDB dataset down to 30 waypoints. Gestures that contained less than 30 points were not considered. This resulted in 6289 sets of waypoints.

For every set $s$ of 30 waypoints in the ILGDB dataset, we calculated the minimum snap trajectories using the BGD (backtracking gradient descent) algorithm, implemented using \cite{wang2021generating}, to obtain the locally optimal cost $J_{BGD}(s)$. The TVP method from \cite{liu2016high} predicts a time allocation for a given set of waypoints, $s$, by using a trapezoidal velocity profile and we use this method to initialize the time allocations for our BGD algorithm. For the TVP method we use a maximum velocity of $5m/s$ and a maximum acceleration of $2.5m/s^2$. In our method, for any given set of points $s$ we obtain the predicted time allocations $\underline{t}$ from our trained transformer model. We then solve Optimization problem \ref{prb: min_snap2} using the time allocations predicted by the transformer to obtain a trajectory with cost $J_T(s)$.


We compare our trained transformer model with prior work in literature, namely \cite{de2019real} and the TVP, \cite{liu2016high}, method used for initialization. In \cite{de2019real} they use an MLP (multi-layer perceptron) that can only be trained for a fixed input-output size. Hence, they train multiple MLPs for different input sizes (3 to 30) to learn the time allocations. The combined models with all 28 MLPs contain $\sim 438,800$ trainable hyper-parameters. The cost/snap of the trajectories generated using the time allocations predicted from the MLP is given by $J_{MLP}$. For fair comparison, all the costs $J_{BGD}, J_T, J_{MLP}$ and $J_{TVP} $ are calculated using the same final time $\sum t_i = T$ obtained using the TVP method. 

We evaluate the optimality of the predicted time allocations from the transformer model based on the total trajectory cost, i.e. total snap. Since the trajectory is a 7th order polynomial (see  \cite{mellinger2011minimum} for further discussion on polynomial order), the total snap of a given trajectory scales with the 7th order of total time $T$, i.e. $f(\alpha \underline{t}) \sim \alpha^7 f(\underline{t})\ \forall\ \alpha>0$ where $f(\underline{t})$ is the snap of the trajectory with time allocations $\underline{t}$. Due to this we normalize the costs by their 7th root, i.e.:
\begin{equation}
   \begin{split}
       &J'_{BGD}(s)= J_{BGD}^{1/7}(s);\ J'_T(s)= J_T^{1/7}(s); \ 
       J'_{MLP}(s)= J_{MLP}^{1/7}(s);\ J'_{TVP}(s) = J_{TVP}^{1/7}(s)
   \end{split} 
\end{equation}

We use the cost/snap of the trajectory generated from BGD method as our baseline. We compare the relative error ($E(s)$) between trajectories generated from other methods and the BGD method to evaluate their optimality. For a given set of waypoints, $s$, we define the relative errors for the different methods, i.e. our method (transformers), MLP method \cite{de2019real} and TVP \cite{liu2016high} as shown in \eqref{eq:E_TVP}

\begin{align}
    \begin{split}
        & E_T(s) = \frac{(J'_T(s) - J'_{BGD}(s))}{J'_{BGD}(s))}\cdot 100; 
    \quad \quad E_{MLP}(s) = \frac{(J'_{MLP}(s) - J'_{BGD}(s))}{J'_{BGD}(s)}\cdot 100; \\
    & E_{TVP}(s) = \frac{(J'_{TVP}(s) - J'_{BGD}(s))}{J'_{BGD}(s)}\cdot 100 
    \end{split}\label{eq:E_TVP}    
\end{align}

\begin{figure}[h]
    \centering
    \includegraphics[width=3.75in]{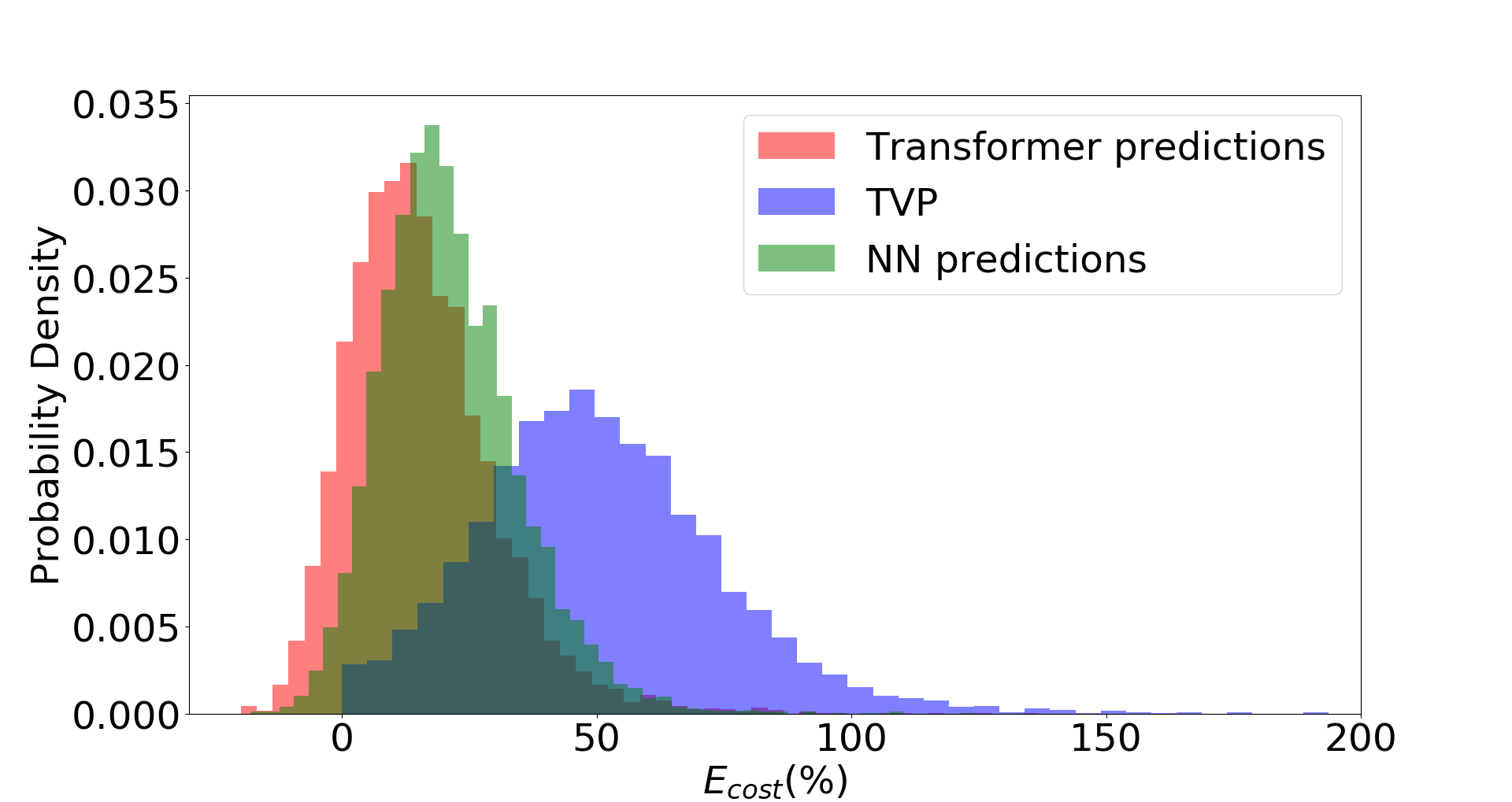}
    \caption{Histograms of error in cost $E_{cost}$ between using transformer and a MLP }
    \label{fig:cost_30}
\end{figure}

Fig \ref{fig:cost_30} shows the histogram of the relative errors, $E(s)$ of the different methods over all sets of waypoints in the ILGDB test dataset. On average, the cost error in the transformer model is $E_{T}^{avg} = 15.7\%$ with standard deviation of $14.6\%$ over the BGD baseline. The MLP method \cite{de2019real} has a mean error in cost of $E_{MLP}^{avg} = 21.4\%$ with standard deviation of $13.7\%$. The TVP initialization had a mean cost error of $E_{TVP}^{avg} = 50.7\%$. 
The figure shows that the time allocations predicted by the transformer model are better than the time allocations predicted by the MLP method. This is because the transformer model is able to use data over all input sizes and model the sequential relation across the input waypoints. We observe that the transformer model is able to guess better time allocations than the baseline BGD method itself. This is represented by the region of the histogram (Fig \ref{fig:cost_30}) with $E_T<0$ and corresponds to $\sim 10.7\%$ of test dataset. The model is able to predict better because the BGD method can get stuck at local minimas leading to sub-optimal time allocations. This was also observed in the MLP model which was able to make better predictions than the baseline ($E_{MLP}<0$) and corresponds to $\sim 3.1\%$ of the test dataset.

\begin{figure}[!h]
    \centering
    \includegraphics[width=4in]{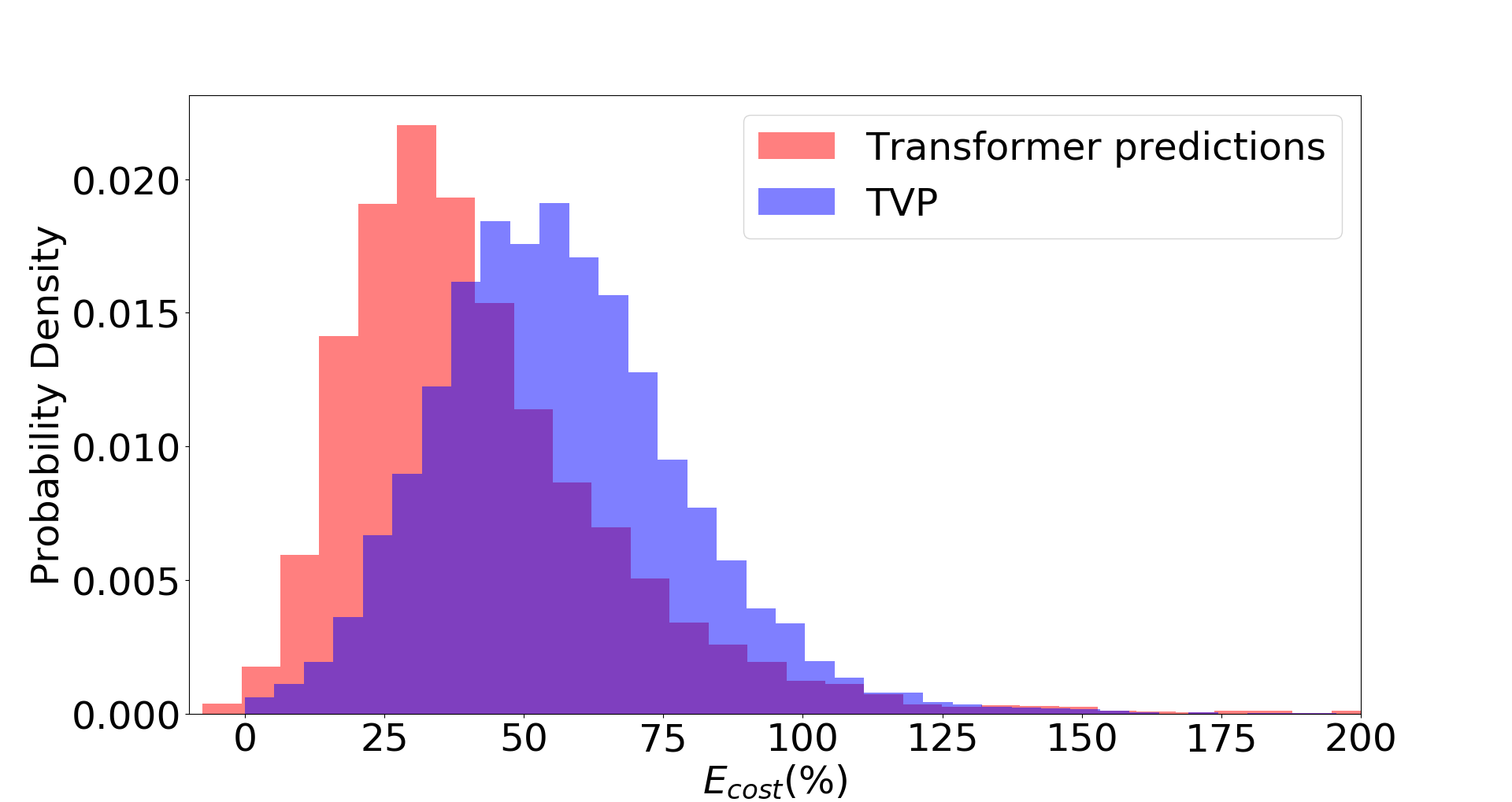}
    \caption{Error histogram in cost $E_{cost}$ from transformer predictions on trajectories with 40 waypoints (out of distribution)}
    \label{fig:cost_40}
\end{figure}

To demonstrate the ability of our approach to model the sequential nature of the problem, we test the transformer model on input sizes outside its training dataset. The gestures in the training dataset were collocated to sizes within $[3,30]$. So we generate a test set by collocating all gestures in the ILGDB dataset to 40 waypoints. Fig \ref{fig:cost_40} shows the error distribution between the predictions of the trained model and their $BGD$ baselines for trajectories with 40 waypoints. This result shows that transformer models can generalize to input sizes larger than the training dataset. As seen in Fig \ref{fig:cost_40}, the transformer predictions are close to the costs obtained by using the BGD method. The average cost error in model predictions for 40 waypoints is $E_{T}^{avg} = 42.7\%$ over the BGD baseline.

All the above trajectory computations had an average computation time in the order of $10^{-3}s$ (Intel-9750H processor and NVIDIA RTX 3060 graphics processor). This however increases quadratically with number of inputs. For real time inference over large inputs ($>10^2)$, we must reduce the context window size of the transformer and use linear self-attention functions such as \cite{wang2020linformer}. Additionally, the attention computation across all context windows can be parallelized to make it faster. These introduce more model parameters that require careful selection. These selection methods have been studied extensively in the literature, such as in speech recognition \cite{sukhbaatar2019adaptive} where thousands of audio packets are processed per second. These modifications are not performed in our work and we only evaluate the performance on the standard transformer.

\subsection{Attention patterns}
\label{ssec:attention_patterns}

\begin{figure}[h]
    \centering
    \begin{subfigure}
        \centering
        \includegraphics[width=2.5in]{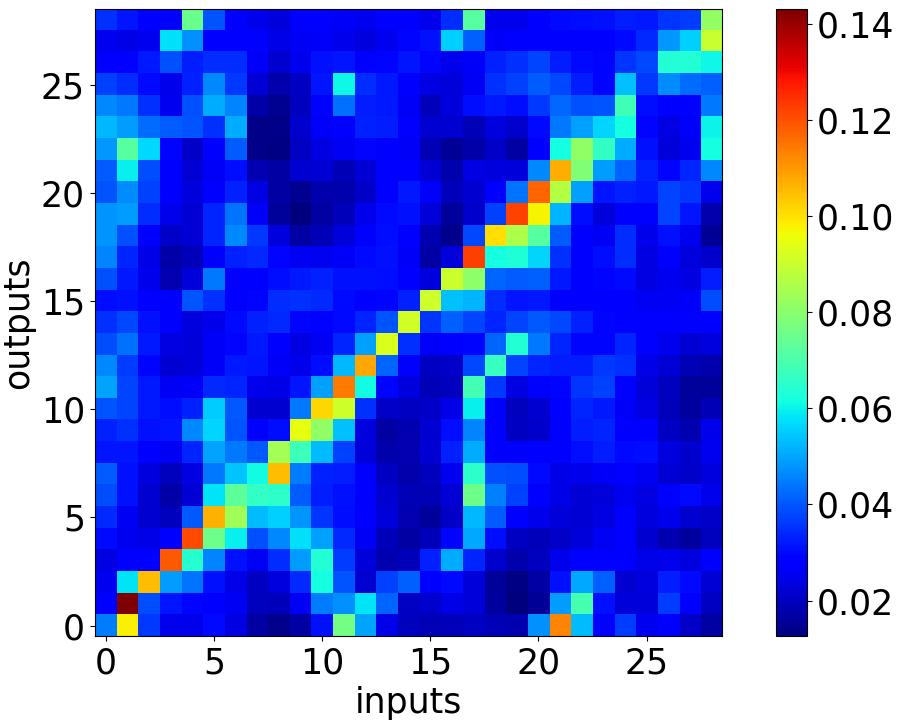}
        \label{fig:attention_layer0}
    \end{subfigure}%
    \begin{subfigure}
        \centering
        \includegraphics[width=2.5in]{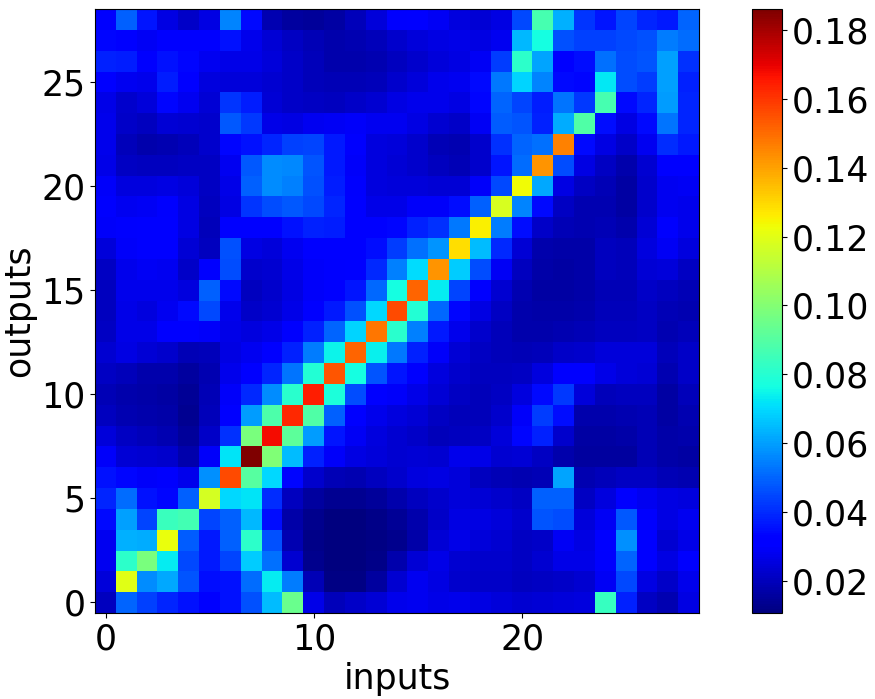}
        \label{fig:attention_layer1}
    \end{subfigure}
    \caption{Averaged attention maps across all heads of encoder-decoder attention block (a) average attention across all heads in layer 1 (b) average attention in layer 2 of encoder-decoder block}
    \label{fig:attention_maps}
\end{figure}

The mean of the attention maps(across all heads) in layer 1 and layer 2 of the encoder-decoder attention block during inference is shown in Fig \ref{fig:attention_maps}. The encoder-decoder attention maps have learnt to attend to the neighbouring prior and future input waypoints. The block diagonal structure indicates that there is a ``horizon'' over which input waypoints influence the time allocation of a given segment. Learning the influence of adjacent waypoints, including the future waypoints helps the transformer make better predictions on the optimal time allocation. Given the block diagonal nature of the attention maps, we can choose a smaller size for the context window. However, we have not used context windows to limit the attention span in this work. 

\subsection{Sample efficiency}
\label{ssec:data_eff}

\begin{figure}[h]
    \centering
    \includegraphics[width=3.5in]{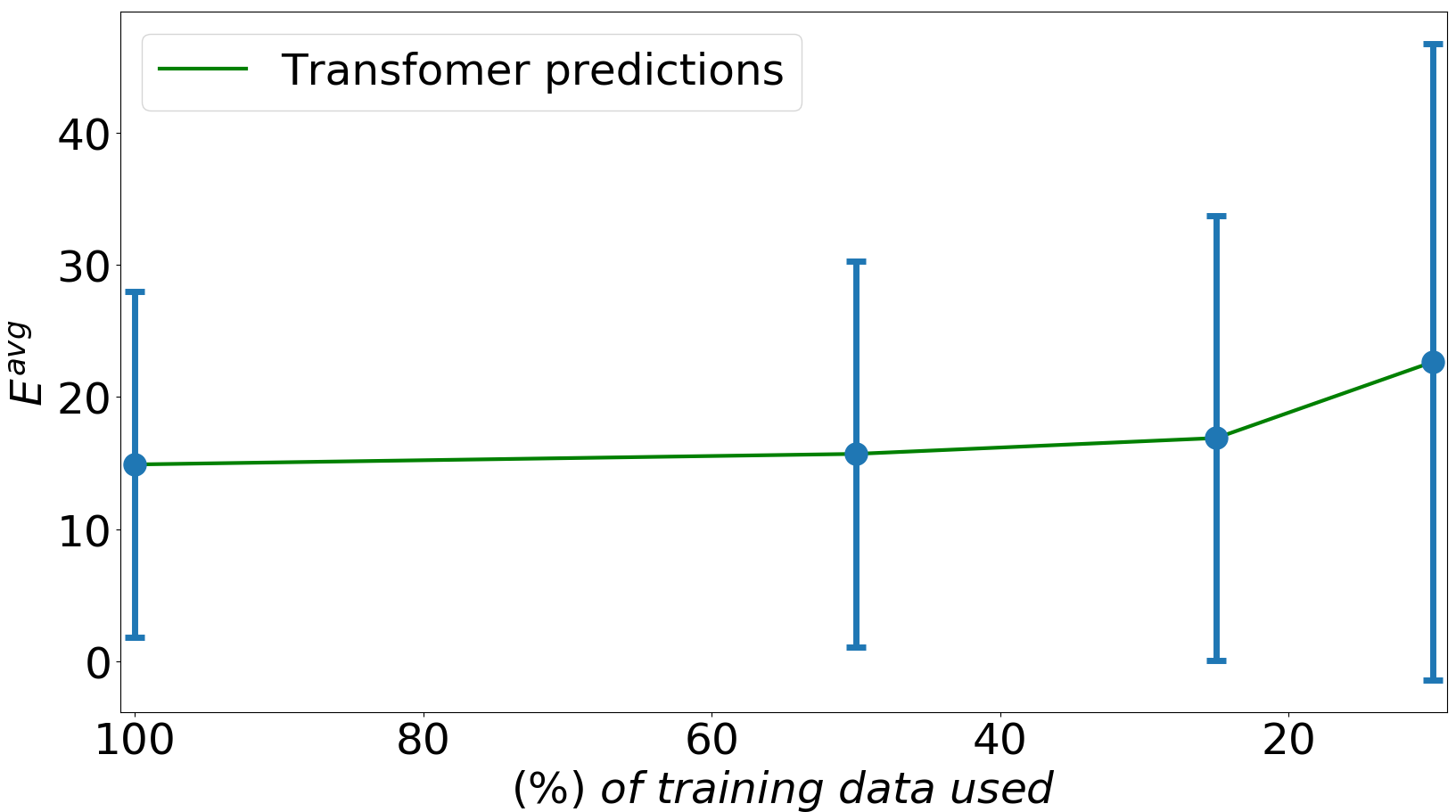}
    \caption{Average cost error over test data vs amount of training data used}
    \label{fig:training_efficiency}
\end{figure}

We study the sensitivity of the model performance with the size of the training dataset. We select random subsets of the training data and measure the performance of the corresponding trained model ($E_T^{mean}$) on the test data (ILGDB). In Fig \ref{fig:training_efficiency} we see that the transformer model is able to perform just as well well as the MLP model using only $10\%$ of the training data ($E_T^{mean}=22.6\%$). This means that the model can provide good predictions using fewer training samples for a given input size, demonstrating its sample efficiency.

\section{Conclusion}
\label{sec:conclusion}
The experiments shown here demonstrate that a trained transformer can successfully predict near optimal time allocations and even provide good starting guesses for input sizes greater than the training data. The transformer model is able to accomplish this with fewer model hyperparameters ($< 1/3^{rd}$) and training data samples ($1/10^{th}$) compared to using an MLP. The learnt attention patterns between input and output embeddings indicate a ``horizon'' of waypoints (before and after) which influence the time allocation at a given waypoint. 

By learning the optimal time allocations for any given sequence of waypoints, our method removes the bi-level structure of the optimization problem. The transformer model was not trained explicitly on the cost function (snap) and instead only on the time allocations. Hence this training approach can be extended to other bi-level optimization problems (minimum jerk and minimum time \cite{zhou2019robust} for example). The major bottleneck for running the above standard transformer model is computing the self-attention functions in real-time for large inputs. This requires several modifications to the transformer model that are discussed in section \ref{sec:result} and is a direction for future work.

The results from our approach establish that trajectory generation methods can be simplified and made faster by learning the computationally expensive optimizations using a sequential learning model (transformer). %



\section{Acknowledgements}
The authors would like to thank Woodside Energy for sponsoring the research presented in this work. We would also like to thank Prof. Philipp Kr\" ahenb\" uhl and Prof. David Fridovich-Keil from The University of Texas at Austin for help with reviewing this paper.

\bibliography{references}  

\end{document}